\documentclass[11pt,draftcls,onecolumn]{IEEEtran}

\newif\ifpgf
\pgftrue

\usepackage{microtype}
\usepackage{graphicx}
\usepackage{subfigure}
\usepackage{booktabs} 

\usepackage{graphicx}
\ifpgf
\usepackage{pgfplots}
\usetikzlibrary{positioning}
\fi
\usepackage{bm}
\usepackage{stmaryrd}
\newcommand{\mkv}{-\!\!\!\!\minuso\!\!\!\!-}
\usepackage{amsfonts}
\usepackage{amsmath}
\usepackage{mathtools}
\usepackage{amssymb,url}
\usepackage{dsfont}
\usepackage{capt-of}

\newcommand{\be}{\begin{equation}}
\newcommand{\ee}{\end{equation}}

\newtheorem{proposition}{Proposition}
\newtheorem{definition}{Definition} 
\newtheorem{lemma}{Lemma}

\usepackage{placeins}

\ifCLASSINFOpdf
\else
\fi
\begin{document}
%

\title{Learning Anonymized Representations with \\[-3mm]Adversarial Neural Networks}

\author{Cl\'ement Feutry,
        Pablo~Piantanida,
        Yoshua~Bengio,
        and~Pierre~Duhamel
 \IEEEcompsocitemizethanks{
\IEEEcompsocthanksitem  C. Feutry, P. Piantanida and P.~Duhamel are with Laboratoire des Signaux et Syst\`emes (L2S, UMR8506), CentraleSup\'elec-CNRS-Universit\'e Paris-Sud, Gif-sur-Yvette, France. Email: \{clement.feutry;pablo.piantanida;pierre.duhamel\}@l2s.centralesupelec.fr.}
\IEEEcompsocitemizethanks{\IEEEcompsocthanksitem Y. Bengio is with Montreal Institute for Learning Algorithms, Universit\'e de Montr\'eal, Montr\'eal, QC, Canada. }
}

\IEEEtitleabstractindextext{%
\begin{abstract}
 Statistical methods protecting sensitive information or the identity of the data owner have become critical to ensure privacy of individuals as well as of organizations. This paper investigates anonymization methods based on representation learning and deep neural networks, and motivated by novel information-theoretical bounds. We introduce a novel training objective for simultaneously training a predictor over target variables of interest (the regular labels) while preventing an intermediate representation to be predictive of the private labels. The architecture is based on three sub-networks: one going from input to representation, one from representation to predicted regular labels, and one from representation to predicted private labels. The training procedure aims at learning representations that preserve the relevant part of the information (about regular labels) while dismissing information about the private labels which correspond to the identity of a person. We demonstrate the success of this approach for two distinct classification versus anonymization tasks (handwritten digits and sentiment analysis). 
\end{abstract}

}
\maketitle


\IEEEdisplaynontitleabstractindextext

%
\IEEEpeerreviewmaketitle
\allowdisplaybreaks

\section{Introduction}
 
In recent years, many datasets containing sensitive information about individuals have been released into public domain with the goal of facilitating data mining research.  Databases are frequently anonymized by simply suppressing identifiers that reveal the identities of the users, like names or identity numbers. However, even these definitions cannot prevent background attacks, in which the attackers already know something about the information contained in the dataset. A popular approach known as differential privacy \cite{10.1007/11787006_1} offers provable privacy guarantees. Intuitively, it uses random noise to ensure that the mechanism outputting information about an underlying dataset is robust to any change of one sample, thus protecting privacy.



In this paper we address the interplay between deep neural networks and statistical  anonymization of datasets. We focus on the following fundamental questions: \emph{What conditions can we place to learn anonymized (or sanitized) representations of a dataset in order to minimize the amount of information which could be revealed about the identity of a person? What is the effect of sanitization on these procedures?}
The line of research we investigate  is based on privacy-preserving statistical methods, such as learning differentially private algorithms~\cite{45428}.  The main goal of this framework  is to enable an analyst to learn relevant properties (e.g., regular labels)  of a dataset as a whole while protecting the privacy of the individual contributors (private labels which can identify a person).  This assumes the database is held by a trusted person who can release freely information about regular labels, e.g., in response to a sequence of queries,  and used for many new purposes.

\subsection{Related work}

The literature in statistics and computer science on anonymization and privacy is extensive; we discuss only directly relevant work here (see~\cite{DBS-008} and references therein). The k-anonymity framework has been introduced by~\cite{sweeney2002k} with the  purpose of processing  databases where each entry is a different person, and each person of the database is described through many features. Several other frameworks linked to k-anonymity such as l-diversity in \cite{machanavajjhala2006diversity} and t-closeness in \cite{li2007t} have been developed a few years later. The main similarity between our framework and k-anonymity is that we do not consider any background knowledge like in k-anonymity. However, the fundamental differences rely on our statistical treatment of the anonymization problem and instead of having only one version of each attribute (or label), we require multiple statistical versions of the same attribute for each individual. Additionally,  databases with k-anonymity contain data that clearly identifies a person whereas we consider datasets where identification can be learned, so we look for data transformations which discard identifying features from the data.

A major challenge in addressing privacy guarantees is ot determine and control the balance between statistical efficiency and the level of privacy, which requires itself a careful mathematical but also meaningful definition.  Typically, these techniques depend on how the data are released  and the literature contains various approaches to this vast problem. The notion of differential privacy has been successfully introduced and largely studied in the literature~\cite{Dwork2008}.  From a statistical perspective, convergence rates for minimax risk for problems  in which the data must be kept  confidential even from the learner have been reported in~\cite{DBLP:journals/corr/abs-0809-4794} and ~\cite{Duchi:2014:PAL:2700084.2666468}. In the machine learning literature, \cite{doi:10.1198/jasa.2009.tm08651} and \cite{Chaudhuri:2011:DPE:1953048.2021036} develop differentially private empirical risk minimization algorithms, and \cite{6979031} and \cite{JMLR:v17:15-313} study similar statistical and sample complexity of differentially private  procedures. \cite{chen2009privacy} and \cite{yuan2014privacy} presented a  privacy-preserving distributed algorithm of backpropagation which allows a neural network to be trained without requiring either party to reveal her data to the other.  \cite{45428} studied differential privacy based on deep neural nets where  each adjacent databases is a set of image-label pairs that differs in a single entry, that is, if one image-label pair is present in one set and absent in the other.


\subsection{Contributions}

We investigate anonymization from a perspective which is related but different from that of differential privacy. The main difference relies on the condition on the information release (sanitize) mechanism which in our case depends on the dataset itself. Additionally, differential privacy introduces randomized predictors whereas our method (after training is accomplished) induces a deterministic algorithm. We do not provide a privacy level of the dataset or of a method. Instead we try to hide information about the private labels which is implicitly present in a dataset while preserving as much information as possible  about the regular relevant labels involved. For this purpose, we introduce a novel training objective and framework inspired by Generative Adversarial Networks (GAN) by~\cite{NIPS2014_5423}  and by the domain adaptation framework of~\cite{DBLP:conf/icml/GaninL15}. We propose an efficient way of optimizing an  information-theoretic objective by deriving backpropagation signals through a competitive process involving three networks, illustrated in Figure~\ref{fig:1}: an encoder network which is a common trunk mapping input $X$ to a representation $U$, as well as two branch networks taking $U$ as input, i.e. a predictor for the regular labels $Y$ and a predictor for the private labels $Z$.
While the encoder is trained to help the predictor of $Y$ as much as possible, it is also trained to prevent the $Z$ predictor from extracting private information from $U$, leading to a trade-off between these two objectives.

This architecture is similar to that of~\cite{DBLP:conf/icml/GaninL15}, initially introduced in the context of domain adaptation. The goal of domain adaptation is to train a loss on a dataset and be able to apply it efficiently on a different but related dataset. Our contributions on top of this architecture are the following. First, we introduce a series of mathematical results based on information-theoretical considerations. Second, they motivate a novel training objective which differs from that of~\cite{DBLP:conf/icml/GaninL15} in two main ways: (a) the adversarial network tries to classify among a large number of person's identities (instead of among two domains), and (b) the training objective is designed to lead to more robust training, avoiding the numerical difficulties which arise if the adversarial cost only tries to increase the cross-entropy of the private-labels predictor. These numerical difficulties arise in particular because minus the cross-entropy (of the private-labels predictor) does not have lower bound, which can lead to very large gradients. A key insight to fix this problem is that such poor behavior happens when the cross-entropy is actually worse than if the private-label predictor was simply producing a uniform distribution over the person's identities, and there is no need to make that predictor have a cross-entropy which is worse than a random guessing predictor.

\subsection*{Notation and conventions} 
Upper-case letters denotes random variables (RVs) and lower-case letters realizations.  
$ \mathbb{E}_{P}[\cdot]$ denotes the expectation w.r.t. $P$ the probability distribution (PD). Let $\mathcal{P}(\mathcal{X})$ denote the set of all PDs in  $\mathcal{X}$.  All empirical PDs computed from samples are denoted by $\hat{P}_{X}$. $\mathbf{P}_X$  is the vector length that contains the values of $P_X$. $|\cdot|$ is used for the usual absolute value and cardinality of a set, and with $\langle\cdot,\cdot\rangle$ the canonical inner product.
All logarithms are taken with base $e$. The information measures are~\cite{Csiszar:1982:ITC:601016}: \emph{entropy}  $ \mathcal{H}(P_X)\coloneqq \mathbb{E}_{P_X}\left[-\log { {P}_X}(X)\right]$;  \emph{conditional  entropy} $ \mathcal{H}(P_{Y|X}|P_X)\coloneqq \mathbb{E}_{P_{X} P_{Y|X}}\left[-\log {P}_{Y|X} (Y|X)\right]$; \emph{mutual information} $\mathcal{I}(P_X;{P}_{Y|X})$; \emph{relative entropy}: $\mathcal{D}( {P}_{X}\| {Q}_{X})$ and \emph{conditional relative entropy}: $\mathcal{D}( {P}_{U|X}\| {Q}_{U|X}| P_X)$.

%

\section{Statistical Model and Problem Definition}

We introduce our model from which sanitized representations will be learned. We develop a precise formalization of the problem and derive an information-theoretic criterion that together GAN  provide a tractable supervised objective to guide the learning of constrained representations.

\subsection{Learning model and problem definition}

In this work, we are concerned with the problem of pattern classification which is about predicting the regular label (public information) of an observation based on high-dimensional representations. An observation is a sample $x\in \mathcal{X}$ presented to the learner about a target concept $y\in \mathcal{Y}$ (the regular label) and the user ID $z\in \mathcal{Z}$ (the private label). This consists of a typical supervised learning setup with a training dataset of $n$ i.i.d. tuples: ${\mathcal{D}_n} \coloneqq  \{(x_1,y_1,z_1)\cdots (x_n,y_n,z_n)\}$, sampled according to an unknown distribution $P_{XYZ}$. We consider learning of a representation from examples of  $P_{XYZ}$. We would like to find a (possibly stochastic) transformation $Q_{U|X}$ that maps raw data $X$ to a higher-dimensional (feature) space $\mathcal{U}$:
\begin{equation*}
P_{YZ} \sim (Y,Z) \xrightarrow[\text{(unknown)}]{P_{X|YZ}}  X \xrightarrow[\text{(encoder/sanitize)}]{Q_{U|X}} U. 
\end{equation*}
This problem can be divided into that of simultaneously finding a (randomized) deep encoder $Q_{U|X}:\mathcal{X}\rightarrow\mathcal{P}(\mathcal{U})$ and a soft-classifier $Q_{\hat{Y}|U}:\mathcal{U}\rightarrow\mathcal{P}(\mathcal{Y})$ which maps the representation to a distribution on the label space $\mathcal{Y}$. Our  ultimate goal is to learn $Q_{U|X}$ from a deep neural network to perform this classification task while preventing any classifier $Q_{\hat{Z}|U}:\mathcal{U}\rightarrow\mathcal{P}(\mathcal{Z})$ from learning the private label $Z$ from the representation $U$. In other words, our representation model must learn invariant features with respect to private labels. We will formalize our problem as being equivalent to that of optimizing a trade-off between the misclassification probabilities so it would be convenient to precisely define this notion: 
\begin{definition} The probability of misclassification of the induced decision rule from an encoder $Q_{U|X}$ and a classifier  $Q_{\hat{Y}|U}$ with respect to the distribution $P_{XY}$ is given by
$$
\hspace{2mm}P_{\mathcal{E}}\big(Q_{U|X}, Q_{\hat{Y}|U} \big) \coloneqq 1- \mathbb{E}_{ P_{XY}Q_{U|X}} \left[ Q_{\hat{Y}|U} (Y|U) \right].
$$
\end{definition}

An upper bound will be used to rewrite this intractable objective into the \emph{cross-entropy  risk} defined below: 
\begin{definition}[Cross-entropy loss]\label{def-Logarithmic-loss}
Given two distributions $Q_{U|X}:\mathcal{X}\rightarrow\mathcal{P}(\mathcal{U})$ and $Q_{\hat{Y}|U}:\mathcal{U}\rightarrow \mathcal{P}(\mathcal{Y})$, define the average (over representations) \emph{cross-entropy loss} as:
\begin{align}
\ell\big({Q}_{U|X}(\cdot|x),Q_{\hat{Y}|U}(y | \cdot)\big)& \coloneqq \big\langle {Q}_{U|X}(\cdot|x) , -\log Q_{\hat{Y}|U}(y | \cdot ) \big\rangle\nonumber\\
&=\mathbb{E}_{Q_{U|X=x}} \left[ - \log Q_{\hat{Y}|U}(y | U)\right].  \label{eq-true-loss}
\end{align}
As usual, we shall measure the expected performance of $({Q}_{U|X},Q_{\hat{Y}|U})$ via the \emph{risk}:
\begin{equation*}
 \mathcal{L}(Q_{\hat{Y}|U},Q_{U|X}) \coloneqq  \mathbb{E}_{{P}_{XY}}   \big[ \ell\big({Q}_{U|X}(\cdot|X),Q_{\hat{Y}|U}(Y | \cdot)\big)  \big]. \label{logarithmic-loss-risk}
\end{equation*}
\end{definition}
We can now provide an operational definition of what would make a  good representation $U$ in the anonymization problem. A representation should be useful for minimizing the misclassification probability of the public task of interest with regular labels $Y$ while bounding from below, whatever classifier $Q_{\hat{Z}|U}$ is chosen, the probability of misclassification of the identity $Z$, which is formally introduced below:  
\begin{definition}[Learning with anonymization] \label{def-learning-probelm}
Consider the following constrained pattern classification problem: 
\begin{equation}
\min\limits_{(Q_{U|X}, Q_{\hat{Y}|U})\in\mathcal{F}} \Big\{ P_{\mathcal{E}}\big(Q_{U|X}, Q_{\hat{Y}|U} \big): \min_{Q_{\hat{Z}|U}:\,\mathcal{U}\rightarrow\mathcal{P}(\mathcal{Z})}  P_{\mathcal{E}}\big(Q_{U|X}, Q_{\hat{Z}|U} \big) \geq 1-\varepsilon  \Big\}, \label{def-trade-off}
\end{equation}
for a prescribed probability $ 1/|\mathcal{Z}| \leq  \varepsilon<1$, where  the minimization is over the set of restricted encoders and classifiers $(Q_{U|X}, Q_{\hat{Y}|U})\in\mathcal{F}$ according to a model class $\mathcal{F}$. 
\end{definition} 
The above expression requires representations with $(1-\varepsilon)$-approximate guarantees (over all possible classifiers) w.r.t. the misclassification probability of the private labels. It is not difficult to see that $\varepsilon$ can be replaced by a  suitable positive multiplier $\lambda\equiv \lambda(\varepsilon)$ yielding an equivalent objective: 
\begin{equation}\label{def-trade-off2}
\min \Big\{ P_{\mathcal{E}}\big(Q_{U|X}, Q_{\hat{Y}|U} \big) - \lambda\cdot P_{\mathcal{E}}\big(Q_{U|X}, Q_{\hat{Z}|U}^\star \big) \Big\},
\end{equation}
where $Q_{\hat{Z}|U}^\star$ is the minimizer of $P_{\mathcal{E}}\big(Q_{U|X}, Q_{\hat{Z}|U} \big)$. Evidently, expression~\eqref{def-trade-off2} does not lead to a tractable objective for training $(Q_{U|X}, Q_{\hat{Y}|U})$. However, it suggests a competitive game between two players: an adversary trying to infer the private labels $Z$ from our representations $U$, by minimizing $P_{\mathcal{E}}\big(Q_{U|X}, Q_{\hat{Z}|U} \big)$ over all possible $Q_{\hat{Z}|U}$, and a legitimate learner predicting the regular labels $Y$, by optimizing a classifier $Q_{\hat{Y}|U}$ over a prescribed model class $\mathcal{F}$. We can trade-off these two quantities via the representation (encoder) model $Q_{U|X}$. This key idea will be further developed in the next section through an adversarial framework to guide learning of all involved parameters in the class $\mathcal{F}$. 
  
\subsection{Bounds on the  probability of misclassification} 

In order to derive a tractable surrogate to \eqref{def-trade-off}, e.g., by relating the probabilities of misclassification   to the corresponding cross-entropy losses,  it is convenient to first introduce the rate-distortion function~\cite{Cover91}. 
\begin{definition}\label{def-rate-distortion}
The rate-distortion function of a RV $Z\in\mathcal{Z}$ with distortion $d(z,u)\coloneqq 1- Q_{\hat{Z}|U}(z|u)$ is defined as:
\begin{equation*}
\mathcal{R}_{Z,Q_{\hat{Z}|U}}(D) \,\coloneqq  \min_{\rule{0mm}{4.3mm}\substack{P_{\hat{U}|Z}:\, \mathcal{Z}\,\rightarrow\mathcal{P}(\mathcal{U}) \\  \mathbb{E}_{P_{\hat{U}Z} }[1- Q_{\hat{Z}|U}(Z|U)] \, \leq\, D}} \!\! \mathcal{I}\big(P_Z;P_{\hat{U}|Z}\big),\label{eq-rate-distortion}
\end{equation*}
where $P_{\hat{U}Z}=P_{\hat{U}|Z}P_Z$. Furthermore, there exists $D>0$ s.t. $\mathcal{R}_{Z,Q_{\hat{Z}|U}}(D)$ is finite~\cite{csiszar74},  let the minimum be $D_{\min}$ with $\displaystyle R_{\max}\coloneqq  \mathcal{R}_{Z,Q_{\hat{Z}|U}}(D)$ as ${D\to D_{\min}+}$. 
\end{definition}
Moreover, it is easy to show that $\mathcal{R}_{Z,Q_{\hat{Z}|U}}(D)$ is positive, monotonically decreasing and convex. Let us define:
\begin{equation*}
\mathcal{R}_{Z,Q_{\hat{Z}|U}}^{-1}(I)\,\coloneqq  \inf\big\{D\in\mathbb{R}_{\geq 0}: \mathcal{R}_{Z,Q_{\hat{Z}|U}}(D)\leq I \big\}
\end{equation*}
which is known as the \emph{distortion-rate} function.
The function $I\mapsto \mathcal{R}_{Z,Q_{\hat{Z}|U}}^{-1}(I)$ is positive and monotonically decreasing.  The  following lemma provides bounds on the misclassification probability  via mutual information and the cross-entropy loss (proof available as supplementary material).
\begin{lemma}\label{lemma-surogate-representations}
The probabilities of misclassification $P_{\mathcal{E}}(Q_{\hat{Y}|U},$ $Q_{{U}|X})$ and $P_{\mathcal{E}}(Q_{\hat{Z}|U},Q_{{U}|X})$ induced by an encoder $Q_{U|X}:\mathcal{X}\rightarrow\mathcal{P}(\mathcal{U})$ and two arbitrary classifiers  $Q_{\hat{Y}|U}:\mathcal{U}\rightarrow\mathcal{P}(\mathcal{Y})$ and $Q_{\hat{Z}|U}:\mathcal{U}\rightarrow\mathcal{P}(\mathcal{Z})$ are bounded by
\begin{align}
 P_{\mathcal{E}}(Q_{\hat{Z}|U},Q_{{U}|X}) & \geq   \mathcal{R}_{Z,Q_{\hat{Z}|U}}^{-1}\left( \mathcal{I}(P_Z;Q_{U|Z})  \right) , \label{eq-surrogate-bound1A}\\
P_{\mathcal{E}}(Q_{\hat{Y}|U},Q_{{U}|Y}) &\leq  1-\exp\left(-\mathcal{L}(Q_{\hat{Y}|U},Q_{U|X})\right),\label{eq-surrogate-bound2}
\end{align}
where $Q_{U|Z}(u|z)=\sum_{x\in\mathcal{X}}Q_{U|X}(u|x)P_{X|Z}(x|z)$.
\end{lemma}
Observe that the lower bound in~\eqref{eq-surrogate-bound1A} is a monotonically decreasing function of the mutual information $\mathcal{I}(P_Z;Q_{U|Z})$. This implies that any limitation of the mutual information between private labels $Z$ and representations $U$ will bound from below the probability of misclassification of private labels,  whatever classifier $Q_{\hat{Z}|U}$ is chosen. On the other hand, the upper bound in~\eqref{eq-surrogate-bound2}  shows that the cross-entropy loss $\mathcal{L}(Q_{\hat{Y}|U},Q_{{U}|X})$ can be used as a surrogate to optimize the misclassification probability of regular labels, which motivates the cross-entropy loss. The practical relevance of these information-theoretic bounds is to provide a mathematical objective for browsing the trade-off~\eqref{def-trade-off} between all feasible misclassification probabilities $P_{\mathcal{E}}\big(Q_{U|X}, Q_{\hat{Y}|U} \big)$ as a function of the prescribed $(1-\varepsilon)$ probability. Therefore, the learner's goal is to select an encoder $Q_{U|X}$ and a classifier $Q_{\hat{Y}|U}$ by minimizing jointly the risk and the mutual information, leading to tightening of both bounds in Lemma~\ref{lemma-surogate-representations}.


Nevertheless, since ${P}_{XYZ}$ is unknown the learner cannot directly measure neither the risk in \eqref{eq-surrogate-bound2} nor the mutual information in \eqref{eq-surrogate-bound1A}. It is common to measure the agreement of a pair of candidates with a training data set based on the empirical data distribution $\hat{P}_{XYZ}$. This yields an information-theoretic objective, being a surrogate of expression~\eqref{def-trade-off2}:  
\begin{equation}
\min\Big\{ \mathcal{L}_{\textrm{emp}}(Q_{\hat{Y}|U},Q_{{U}|X}) + \lambda\cdot \mathcal{I}(\hat{P}_Z;\hat{Q}_{U|Z})\Big\}, \label{IT-trade-off}
\end{equation}
for a suitable multiplier $\lambda\geq 0$, where $\mathcal{L}_{\textrm{emp}}(Q_{\hat{Y}|U},Q_{{U}|X})$ denotes the 
 \emph{empirical risk} as in Definition~\ref{def-Logarithmic-loss} taking the average w.r.t.  $\hat{P}_{XY}$ and the mutual information must be evaluated using $\hat{Q}_{Z|U}$ as being the posterior according to $Q_{{U}|X} \hat{P}_{XZ}$. As a matter of fact, \eqref{IT-trade-off} may be independently motivated by a rather different problem studying distortion-equivocation trade-offs~\cite{Villard:2013:SMS:2689735.2690352}.

\subsection{Representation learning with anonymization}
 
We performed initial experiments in which the training objective was similar to the one introduced by
~\cite{DBLP:conf/icml/GaninL15} and found that training was unstable and led to a poor trade-off between the degree of anonymity (with the classification error on private labels $Z$ as a proxy) and the accuracy on the regular task (predicting regular labels  $Y$). This led us to change both the training objective and the training procedure, compared to those proposed by ~\cite{DBLP:conf/icml/GaninL15}. 
The new adversarial training objective is presented below, starting from the information-theoretic surrogate presented above in expression \eqref{IT-trade-off}.

A careful examination of expression~\eqref{IT-trade-off} shows that it cannot be optimized since the posterior distribution $\hat{Q}_{Z|U}$ is still not computable in high dimensions. We will further looser this surrogate by upper bounding the mutual information $\mathcal{I}(\hat{P}_Z;\hat{Q}_{U|Z}) =  \mathcal{H}(\hat{P}_{Z}) - \mathcal{H}(\hat{Q}_{Z|U}|\hat{Q}_U)$. The \emph{empirical entropy} of $Z$ can be upper bounded as follows:  
\begin{align}
\mathcal{H}(\hat{P}_Z) &\leq  \mathbb{E}_{\hat{P}_{Z}}   \big[ - \log  \hat{Q}_{\hat{Z}}(Z) \big]  \label{eq-approximationC}\\
&\leq  \mathbb{E}_{\hat{P}_{Z}}  \mathbb{E}_{\hat{Q}_{U}}   \big[ - \log  Q_{\hat{Z}|U}(Z|U) \big]   \label{eq-approximationB}\\
&\equiv \mathbb{E}_{\hat{P}_{Z}} \mathbb{E}_{\hat{P}_{X}}   \big[ \ell\big({Q}_{U|X}(\cdot|X),Q_{\hat{Z}|U}(Z | \cdot)\big)  \big]  \label{eq-approximationAA}\\
&\coloneqq   \mathcal{L}_{\textrm{emp}}^{\textrm{obj}}(Q_{\hat{Z}|U},Q_{{U}|X}), \label{eq-approximationA}
\end{align} 
where \eqref{eq-approximationC} follows since the relative entropy is non-negative; \eqref{eq-approximationB} follows by the convexity of $t\mapsto -\log(t)$ and  \eqref{eq-approximationAA} follows from the definition of the cross-entropy loss. We will also resort to an approximation of the conditional entropy $\mathcal{H}(\hat{Q}_{Z|U}|\hat{Q}_U)$ by an adequate empirical cross-entropy risk:
\begin{align}
\mathcal{H}(\hat{Q}_{Z|U}|\hat{Q}_U) &\approx \mathbb{E}_{\hat{P}_{XZ}}   \big[ \ell\big({Q}_{U|X}(\cdot|X),Q_{\hat{Z}|U}(Z | \cdot)\big)  \big] ,\nonumber\\
&\equiv \mathcal{L}_{\textrm{emp}}(Q_{\hat{Z}|U},Q_{{U}|X}) \label{eq-approximation}
\end{align} 
which assumes a well-selected classifier $Q_{\hat{Z}|U}$, i.e., the resulting approximation error $\mathcal{D}\big(\hat{Q}_{Z|U}\|Q_{\hat{Z}|U}|\hat{Q}_U\big)$ w.r.t. the exact $Q_{\hat{Z}|U}$ is small enough. By combining expressions~\eqref{eq-approximationA} and~\eqref{eq-approximation}, and taking the absolute value, we obtain:
\begin{equation*}  
\mathcal{I}(\hat{P}_Z;\hat{Q}_{U|Z})  \precsim \left|  \mathcal{L}_{\textrm{emp}}^{\textrm{obj}}(Q_{\hat{Z}|U},Q_{{U}|X}) - \mathcal{L}_{\textrm{emp}}(Q_{\hat{Z}|U},Q_{{U}|X})\right| 
\end{equation*}
that together with \eqref{IT-trade-off} leads to our tractable objective for learning, which is an approximation of expression \eqref{IT-trade-off}, being the surrogate of~\eqref{def-trade-off2}, i.e., the objective of interest:  
\begin{align}
\mathcal{L}_{\lambda}(Q_{\hat{Y}|U},Q_{\hat{Z}|U},Q_{{U}|X}) &\coloneqq  \mathcal{L}_{\textrm{emp}}(Q_{\hat{Y}|U},Q_{{U}|X})\nonumber\\ 
&\hspace{1mm}+ \lambda \cdot\left|  \mathcal{L}_{\textrm{emp}}^{\textrm{obj}}(Q_{\hat{Z}|U},Q_{{U}|X}) - \mathcal{L}_{\textrm{emp}}(Q_{\hat{Z}|U},Q_{{U}|X})\right| , \label{IT-trade-off3}
\end{align}
for a suitable  classifier $Q_{\hat{Z}|U}$ and multiplier $\lambda\geq 0$, being a meta-parameter that controls the sensitive trade-off between data anonymity and statistical efficiency. Consequently, we can minimize and maximize the incompatible objectives of the \emph{cross-entropy losses} in \eqref{IT-trade-off3}. Intuitively, the data representations we wish to achieve from ${Q}_{U|X}$ must blur the private labels $Z$ from the raw data $X$ while preserving as much as possible relevant information about the regular labels $Y$. It is worth to mention that \eqref{eq-approximationA} corresponds to the loss of a `random guessing' classifier in which the representations $U$ are independent of private labels $Z$. As a consequence, training encoders $Q_{U|X}$ to minimize~\eqref{IT-trade-off3} enforces the best classifier $Q_{\hat{Z}|U}$ (private labels) to get closer --in terms of loss-- to the random guessing classifier. 

\subsection{Estimation of the probability of misclassification}

The following proposition provides an interesting lower bound on the estimated (e.g. over a choice of test-set)  misclassification probability of any classifier attempting to learn $Z$ from the  released representations: 
\begin{proposition}
Let $Q_{U|X}$ be a sanitize encoder and $\hat{P}_{XZ}$ be an empirical distribution over a choice of a data-set  ${\mathcal{D}_n} \coloneqq  \{(x_1,z_1)\cdots (x_n,z_n)\}$. Then, the probability of misclassification of private labels satisfies: 
\begin{align}
\hat{P}_{\mathcal{E}}(Q_{\hat{Z}|U}, Q_{{U}|X}) & \coloneqq 1- \frac1n \sum\limits_{i=1}^n  \mathbb{E}_{ Q_{U|x_i}} \left[ Q_{\hat{Z}|U} (z_i |U) \right] \nonumber\\
& \geq  g^{-1}\left(\log \left| \mathcal{Z} \right| - \mathcal{I}\big(\hat{P}_Z;\hat{Q}_{U|Z} \big) \right) ,  \label{propo-capacity2}
\end{align}
uniformly over the choice of $Q_{\hat{Z}|U}$, where for $0\leq t\leq 1$:   $g(t) \coloneqq t \cdot\log \left(\left| \mathcal{Z} \right| -1\right) + H(t)$ with $H(t) \coloneqq -t\log(t)-(1-t)\log(1-t)$ and $0\log 0 \coloneqq 0$. The function  $g^{-1}(t)\coloneqq 0$ for $t<0$ and, for $0<t< \log \left| \mathcal{Z} \right|$, $g^{-1}(t)$ is a solution of the equation $g(\varepsilon)=t$ w.r.t. $\varepsilon\in\big[0,  1-1/\left| \mathcal{Z} \right| \big]$; this solution exists since the function $g$ is continuous and increasing on  $\big[0, 1-1/\left| \mathcal{Z} \right|\big]$ and $g(0)=0$, $g\big(1-1/\left| \mathcal{Z} \right| \big) =  \log \left| \mathcal{Z} \right|$. 
\end{proposition}
The proof of this proposition follows by applying Lemma 2.10 in~\cite{Tsybakov:2008:INE:1522486} from which we can bound from below the misclassification probability and will be omitted.  

The importance of expression~\eqref{propo-capacity2} is that it provides a concrete measure for the anonymization performance of the representations. It bounds from below the misclassification probability over the choice of the  classifier $Q_{\hat{Z}|U}$, using the sanitize representations. The right hand side is a quantity that involves the empirical mutual information between the representations and the private labels. It should be pointed out that since in many cases
$\mathcal{H}(\hat{P}_{Z}) \approx \mathcal{H}({P}_{Z}) \equiv \log | \mathcal{Z} | $, assuming ${P}_{Z}$ is uniformly distributed over the set $\mathcal{Z}$, then: 
\begin{equation}
\inf\limits_{Q_{\hat{Z}|U}} \hat{P}_{\mathcal{E}}(Q_{\hat{Z}|U}, Q_{{U}|X})  \succsim g^{-1}\left( \mathcal{H}\big(\hat{Q}_{Z|U} |\hat{Q}_U\big) \right),  \label{propo-capacity3}
\end{equation} 
and using our approximation in~\eqref{eq-approximation} the lower bound in~\eqref{propo-capacity3} leads to an effective and  computable lower bound on the misclassification probability of the private labels.  However, in order to provide statistical  guarantees on \eqref{propo-capacity3}, we need to study confidential bounds on $\mathcal{D}(\hat{Q}_{Z|U}\|Q_{\hat{Z}|U}|\hat{Q}_U)\leq \delta$  which goes  beyond the scope of this paper.

\section{Anonymization with Deep Neural Networks}
Our ultimate goal is to learn parameters $\mathds{R}^{d_c} \ni \bm{\theta}_c\mapsto Q_{U|X}$ of a deep encoder  and  parameters $\mathds{R}^{d_r} \ni \bm{\theta}_r\mapsto  Q_{\hat{Y}|U}$ and $\mathds{R}^{d_p} \ni \bm{\theta}_p\mapsto  Q_{\hat{Z}|U}$ of the classifiers, $(d_c,d_r,d_p)$ being the parameters' dimensions. In the following, we introduce a simplified notation to rewrite the objective~\eqref{IT-trade-off3} as: 
\begin{equation}
\bm{\theta}^*\equiv  \arg\min_{\bm{\theta}\in\Theta } \left\{\mathcal{L}_r(\bm{\theta}_c,\bm{\theta}_r) -\lambda \cdot \left| \mathcal{L}^{\textrm{obj}}_{p}(\bm{\theta}_c,\bm{\theta}_p)- \mathcal{L}_{p}(\bm{\theta}_c,\bm{\theta}_p) \right| \right\}, \label{eq-losses-lagrange}
\end{equation}
for a suitable hyperparameter $\lambda\geq 0 $ to tune the trade-off between regular and private tasks,  where  all  involved parameters are simply denoted by $\Theta \ni \bm{\theta}\coloneqq  (\bm{\theta}_c,\bm{\theta}_r,\bm{\theta}_p)$ with 
\begin{align}
\mathcal{L}_r(\bm{\theta}_c,\bm{\theta}_r) &\equiv \mathcal{L}_{\textrm{emp}}(Q_{\hat{Y}|U},Q_{{U}|X}),\label{eq1-missing}\\ 
\mathcal{L}_{p}(\bm{\theta}_c,\bm{\theta}_p) &\equiv  \mathcal{L}_{\textrm{emp}}(Q_{\hat{Z}|U},Q_{{U}|X}),\label{eq2-missing}\\
\mathcal{L}^{\textrm{obj}}_{p}(\bm{\theta}_c,\bm{\theta}_p) &\equiv \mathcal{L}_{\textrm{emp}}^{\textrm{obj}}(Q_{\hat{Z}|U},Q_{{U}|X}).\label{eq3-missing}
\end{align}
 Assume a training set $\mathcal{D}_n$ of size $n$, where each element of the dataset $(\bm{x}_i,y_i,z_i)$ is composed of $\bm{x}_i\in \mathcal{X} \equiv \mathbb{R}^m$ is a real vector of size $m$, the regular label of the sample $y_i \in \mathcal{Y}$ and  private label of the sample $z_i \in \mathcal{Z}$.

\subsection{Adversarial training objective}

Each classifier branch of the proposed architecture, i.e., $Q_{\hat{Y}|U}$ and $Q_{\hat{Z}|U}$, is trained to minimize the associated cross-entropy loss, whereas the encoder $Q_{U|X}$ will be trained to simultaneously minimize the cross-entropy loss on the prediction of $Y$ while maximizing an adversarial loss defined with respect to the private label predictor $Z$. 

Each sample input $\bm{x}_i$ produces a representation $\bm{u}_i \sim Q_{U|\bm{X}=\bm{x}_i}$ and outputs two probability vectors $\bm{Q}_{\hat{Y}|U}(\cdot|\bm{u}_i)$ and $\bm{Q}_{\hat{Z}|U}(\cdot|\bm{u}_i)$ as soft predictions of the true labels: the regular one $y_i$ and the private one $z_i$, respectively. The expressions of the losses  we found in~\eqref{eq1-missing} and~\eqref{eq2-missing}  are two cross-entropies computed over the whole training set:
\begin{align}
  \label{eq:y-cross-entropy}
  \mathcal{L}_r(\bm{\theta}_c,\bm{\theta}_r) &= \dfrac{1}{n}\sum_{i=1}^{n} \big\langle \bm{e}(y_i),-\log \bm{Q}_{\hat{Y}|U}(\cdot|\bm{u}_i)\big\rangle,\\
\label{eq:z-cross-entropy}  
\mathcal{L}_{p}(\bm{\theta}_c,\bm{\theta}_p) &= \dfrac{1}{n}\sum_{i=1}^{n} \big\langle \bm{e}(z_i),-\log \bm{Q}_{\hat{Z}|U}(\cdot|\bm{u}_i)\big\rangle,
\end{align}
with $\bm{e}(y_i)$ and $\bm{e}(z_i)$ being ``one-hot'' vectors ($y_i$ component is 1 and the others 0) of the true labels of sample $i=[1:n]$.

Let us now consider the adversarial objective.
There are too many possible networks that mismatch the private labels and
maximize the corresponding cross-entropy.
In particular the cross-entropy loss on the private label predictor could be
increased arbitrarily by making it produce a wrong answer with high probability,
which would not make much sense in our context. Hence, we want to maximize this
cross-entropy but not more than that of the cross-entropy of a predictor which would
be unable to distinguish among the identities, i.e., with a posterior distribution approximatly equal to $\hat{P}_Z$:
\begin{equation}
\mathcal{L}_p^{\textrm{obj}}(\bm{\theta}_c,\bm{\theta}_p)=   \dfrac{1}{n}\sum_{i=1}^{n} \big \langle \bm{\hat{P}}_Z, - \log \bm{Q}_{\hat{Z}|U}(\cdot|\bm{u}_i)\big \rangle,\label{eq-obj}
\end{equation}
which is indeed expression~\eqref{eq3-missing}. This artificial loss, formally introduced by our surrogate~\eqref{IT-trade-off3},  denotes the cross-entropy between the vector of  empirical  estimates of probabilities $\bm{\hat{P}}_Z$ and the predictions $\hat{\bm{z}}$. By forcing  private task predictions to follow the estimated probability distribution of the private labels (in many cases close to equiprobable labels) the model output is expected to be as bad as random guessing private labels. Keep in mind that random guessing is a universal lower bound for anonymization.  In fact, if the private label predictor had a cross-entropy loss higher than that of the random guessing predictor, the surrogate indicates we must {\em reduce} its loss. This  is consistent with the adversarial training objective in \eqref{eq-losses-lagrange}. Notice that if our predictions follow the random guessing distribution then the term $\big| \mathcal{L}_p^{obj} (\bm{\theta}_c,\bm{\theta}_p) - \mathcal{L}_{p}(\bm{\theta}_c,\bm{\theta}_p) \big|$ approaches zero.

{\centering
\resizebox{12cm}{!}{%
\def\layersep{1.5cm}
\def\layersepshort{1cm}
  \begin{tikzpicture}[shorten >=1pt,->,draw=black!50, node distance=\layersep]
    \tikzstyle{every pin edge}=[<-,shorten <=1pt]
    \tikzstyle{neuron}=[circle,fill=black!25,minimum size=17pt,inner sep=0pt]
    \tikzstyle{input neuron}=[neuron, fill=none];
    \tikzstyle{output neuron}=[neuron, fill=none];
    \tikzstyle{commun neuron}=[neuron, fill=yellow,draw=black, thick];
    \tikzstyle{classic neuron}=[neuron, fill=green,draw=black, thick];
    \tikzstyle{private neuron}=[neuron, fill=red,draw=black, thick];
    \tikzstyle{hidden neuron}=[neuron,fill=blue,draw=black, thick];
    \tikzstyle{hidden neuron 2}=[neuron,fill=none];
    \tikzstyle{annot} = [text width=4em, text centered]

    \foreach \name / \y in {1,...,3}
        \node[input neuron,] (I-\name) at (0,-\y) {$x_{\y}$};
 \node[classic neuron] (Cllegend) at (-0.6,-4.6 cm) {};
 \node[private neuron] (Cplegend) at (-0.6,-5.8 cm) {};
 \node[input neuron,] (CRlegend1) at (1.6,-5.2 cm) {$\ $};
 \node[input neuron,] (CRlegend2) at (4.6,-5.2 cm) {$\ $};
 \node[input neuron,] (CRlegend3) at (3.1,-5.6 cm) {Gradient reversal layer};
 \path (CRlegend1) edge[purple!50] (CRlegend2);

    \foreach \name / \y in {1,...,4}
        \path[yshift=0.5cm]
            node[hidden neuron] (H-\name) at (\layersep,-\y cm) {};
    \foreach \name / \y in {1,...,4}
        \path[yshift=0.5cm]
            node[hidden neuron] (Hh-\name) at (2*\layersep,-\y cm) {};
	\foreach \name / \y in {1,...,4}
        \path[yshift=0.5cm]
            node[hidden neuron] (Ct-\name) at (3*\layersep,-\y cm) {};
    \foreach \name / \y in {1,...,4}
        \path[yshift=2.6cm]
            node[classic neuron] (Cl-\name) at (4*\layersep,-\y cm) {};
            
            \foreach \name / \y in {1,...,4}
        \path[yshift=-1.6cm]
            node[private neuron] (Cp-\name) at (4*\layersep,-\y cm) {};
            \foreach \name / \y in {1,...,4}
        \path[yshift=2.6cm]
            node[classic neuron] (Cll-\name) at (5*\layersep,-\y cm) {};
            \foreach \name / \y in {1,...,4}
        \path[yshift=-1.6cm]
            node[private neuron] (Cpp-\name) at (5*\layersep,-\y cm) {};
	\foreach \name / \y in {1,...,4}
        \path[yshift=2.6cm]
            node[hidden neuron 2] (Hhl-\name) at (6*\layersep,-\y cm) {};
     \foreach \name / \y in {1,...,4}
        \path[yshift=-1.6cm]
            node[hidden neuron 2] (Hhp-\name) at (6*\layersep,-\y cm) {};
     \foreach \name / \y in {1,...,4}
        \path[yshift=2.6cm]
            node[output neuron] (Ol-\name) at (7*\layersep,-\y cm) {$Q_{\hat{Y}|U}(\y|\bm{u}_i)$};
            \foreach \name / \y in {1,...,4}
        \path[yshift=-1.6cm]
            node[output neuron] (Op-\name) at (7*\layersep,-\y cm) {$Q_{\hat{Z}|U}(\y|\bm{u}_i)$};      
	\draw[black,thick] (8.7,1.8) -- (9.3,1.8) -- (9.3,-1.5) -- (8.7,-1.5) -- cycle;
	\draw (9,0.15) node[rotate=90] {Softmax};
	\draw[black,thick] (8.7,-5.8) -- (9.3,-5.8) -- (9.3,-2.4) -- (8.7,-2.4) -- cycle;
	\draw (9,-4.1) node[rotate=90] {Softmax};
    \foreach \source in {1,...,3}
        \foreach \dest in {1,...,4}
            \path (I-\source) edge (H-\dest);
    \foreach \source in {1,...,4}
        \foreach \dest in {1,...,4}
            \path (H-\source) edge (Hh-\dest);
     \foreach \source in {1,...,4}
        \foreach \dest in {1,...,4}
        \path (Hh-\source) edge (Ct-\dest);
       \foreach \source in {1,...,4}
        \foreach \dest in {1,...,4}
        \path (Ct-\source) edge (Cl-\dest);
       \foreach \source in {1,...,4}
        \foreach \dest in {1,...,4}
        \path (Ct-\source) edge[purple!50] (Cp-\dest);
        \foreach \source in {1,...,4}
        \foreach \dest in {1,...,4}
        \path (Cl-\source) edge (Cll-\dest);
        \foreach \source in {1,...,4}
        \foreach \dest in {1,...,4}
        \path (Cp-\source) edge (Cpp-\dest);
	\foreach \source in {1,...,4}
        \path (Cll-\source) edge (Hhl-\source);
        \foreach \source in {1,...,4}
        \path (Cpp-\source) edge (Hhp-\source);
        \foreach \source in {1,...,4}
        \path (Hhl-\source) edge (Ol-\source);
        \foreach \source in {1,...,4}
        \path (Hhp-\source) edge (Op-\source);
    \node[annot,above of=I-1, node distance=1.45cm] (hl) {$\bm{x}$};
    \node[annot,above of=Hh-1,node distance=1cm] { Encoder};
    \node[annot,right of=Cllegend,node distance=1cm] {regular branch};
    \node[annot,right of=Cplegend,node distance=1cm] {private branch};
    \end{tikzpicture}}
\captionof{figure}{Architecture of the proposed deep neural network.}\label{fig:1}
}

\subsection{Training procedure}
We have found best results according to the following adversarial training procedure, described in Figure~\ref{fig:1}. 
\begin{enumerate}
\item The encoder and regular label predictor are jointly pre-trained (as a standard deep network) to minimize the regular label cross-entropy (eq.~\ref{eq:y-cross-entropy}).
\item The encoder is frozen and the private label predictor is pre-trained to minimize its cross-entropy (eq.~\ref{eq:z-cross-entropy}).
\item Adversarial training is organized by alternatively either training the branch predictors or training the encoder:
  \begin{enumerate}
  \item Sample $N$ training examples and update both branch predictors with respect to their associated cross-entropies, using minibatch SGD (i.e. the $N$ examples are broken down into minibatches, with one update after each minibatch).
  \item Sample $N$ training examples and update the encoder to minimize the adversarial objective (eq.~\ref{eq-losses-lagrange}), again using minibatch SGD.
  \end{enumerate}
\end{enumerate}
In our experiments, we simply picked $N$ as the size of the training set, so we alternated between the two kinds of updates after several epochs on each. We used minibatch SGD with Nesterov momentum~\cite{nesterov2007gradient}.

\section{Experimental Results}

\begin{figure}[t!]
  \centering
\includegraphics[width=15cm]{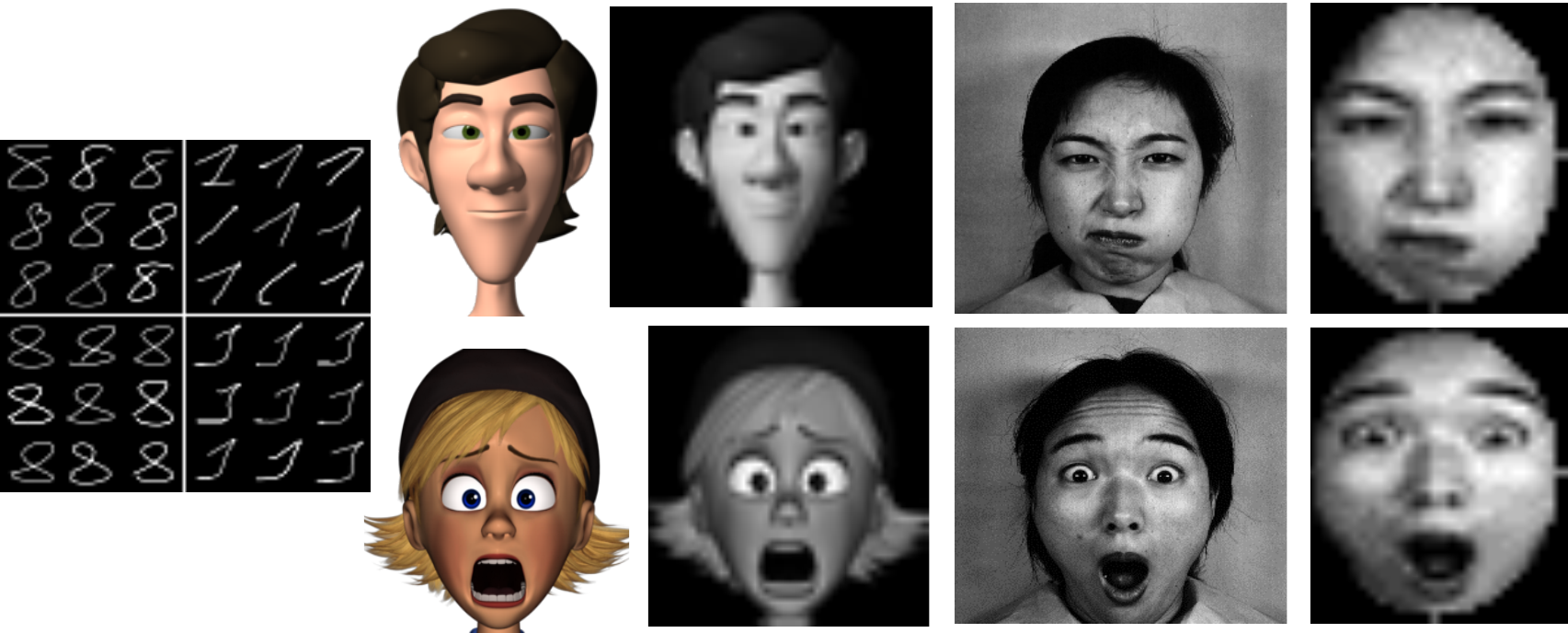} 
  \caption{Samples of preprocessed pen-digits (images on the left), JAFFE (images on the right) and FERG (images at the center).}
  \label{samplependigit}
\end{figure}

\subsection{Presentation of the datasets}

\subsubsection*{Classification of digits (Pen-digits database)} We selected a simple enough dataset, named Pen-digits from Alpaydin~\cite{alimoglu1996methods}. This dataset is interesting to study anonymization because it has double labels (the user IDs of writers and the digit categories) and it has many examples of each writer.
The dataset provides the coordinates of digitally acquired pen movements of 44 persons (30 are involved in the training set and 14 in the test-set)  writing digits from 0 to 9. We only used the training set which was randomly split into training, validation and test data sets (size 5494, 1000 and 1000, respectively), sharing images of the same 30 persons. At the time of collecting this dataset, inconclusive digits were removed. This dataset contains 25 times each digits for each person minus the few discarded digits. The dataset is split in a training part and and a test part.  The raw data is a set of pen trajectories. It is preprocessed in several steps. The coordinates of all the curves corresponding to a single sample were normalized in order to center the image and reduce variability by making it fit a 80x80 image. Each image was then down-sampled into a 20x20 image. The network has 700 neurons per layer and a dropout probability $p_{\textrm{out}}=0.1$ is selected. The encoder is composed of 8 layers and each branch is formed by 3 layers, with all layers except the branch outputs having rectified linear units as non-linearity. The last layer of each branch is a \emph{softmax} output layer.

%

\subsubsection*{Sentiment analysis (FERG database)} The FERG database \cite{aneja2016modeling} contains 55767 annotated face synthetic images of six stylized characters modeled using the MAYA software. This database has 256x256 images depicting  the seven following facial expressions (or feelings): ``neutral'', ``anger'', ``fear'', ``surprise'', ``sadness'', ``joy'' and ``disgust''. For each expression and character, there is between 911 and 2088 images. Original colour images have been pre-processed into a 8-bit grey-scale 50x50 images. The network is composed of 1200 neurones per-layer. The encoder is composed of 5 layers and each branch is formed by 3 layers, other network parameters remain the same as in our  previous network configuration. 

\subsubsection*{Sentiment analysis (JAFFE database)} The JAFFE database \cite{Lyons:1998:CFE:520809.796143} and \cite{dailey2010evidence} contains 213 pictures of Japanese women's faces composed of 10 different persons, where each presents between 2 and 4 pictures per facial expression (of the seven feelings). The pictures were processed to remove irrelevant background pixels. Pictures have been cut in order to have the bottom of the chin as the bottom pixels line, the frontier between hair and forehead as the top pixels line, the frontier between hair and temple as the far right and far left pixels columns. The remaining pixels in the corner that do not belong to the face were set to black. The original pictures are 256x256 pixels and the resulting images are 29x37 pixels. The choice of downsizing the pictures is motivated by the number of samples which is rather small compared to the initial size of the pictures. The dataset is divided into a 139 pictures training set and a 74 pictures test set. There is barely enough data to perform the training properly so the training set is used as the validation set as well. This decision may be considered as fallacious but a validation set is needed because several steps of the algorithm are optimized with the loss value or the accuracy value on the validation set.
 The network used to perform the simulation over this database is a multi-layer perceptron which is not the most efficient one given the small dataset. However, the main purpose of this simulation is to provide a proof of concept for our algorithm. Despite being weak, the overall performance on this recognition task should be sufficient for our purpose. 
\begin{figure}[ht]
  \centering
\includegraphics[width=12cm]{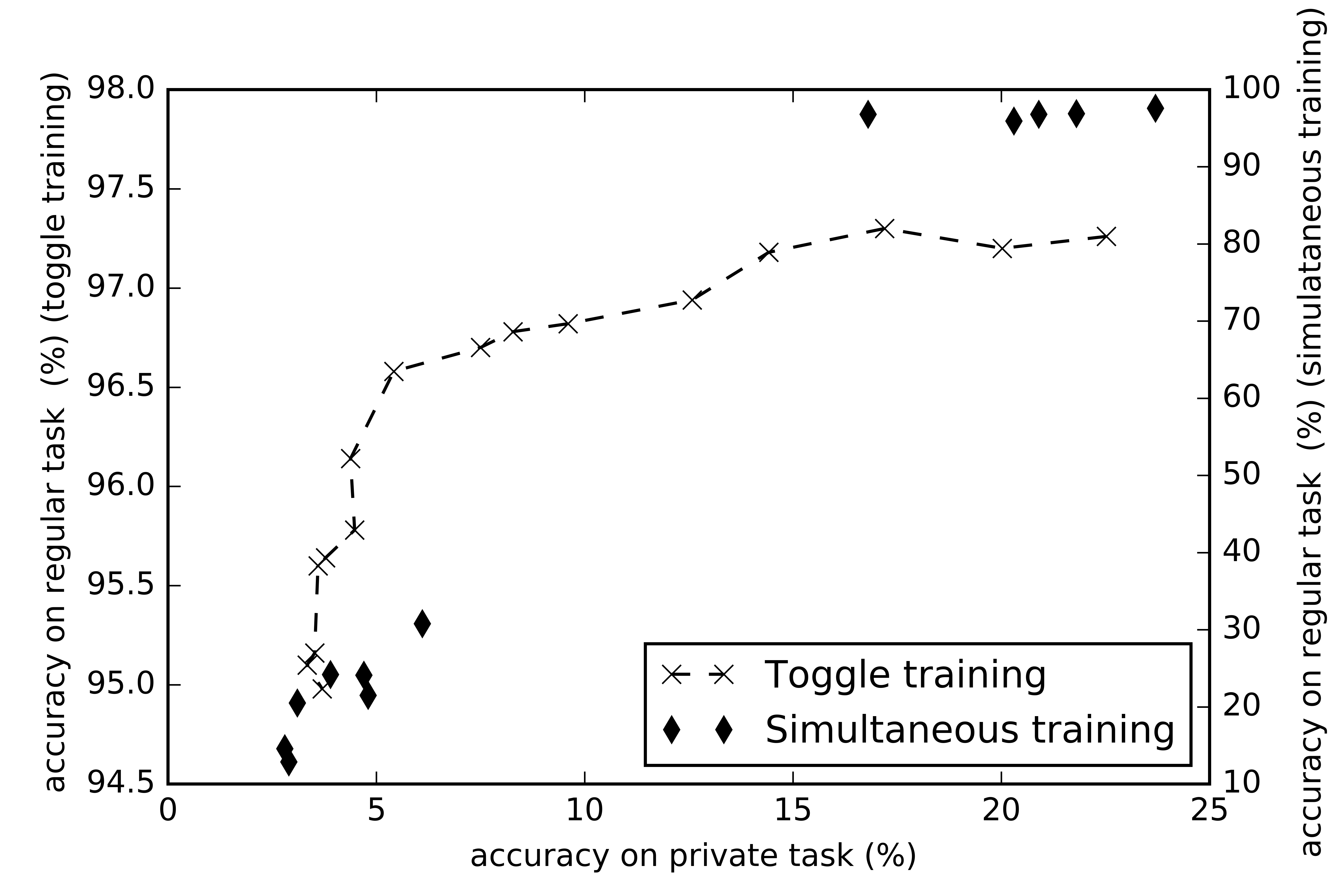} 
\caption{Comparison of the accuracy on regular task between toggle training and simultaneous training, on the Pen-digits database, as a function of the accuracy on the private task. Toggle training provides a better trade-off for the anonymization than the simultaneous training. Simultaneous training enables only two regimes: either a light anonymization, with almost no trade-off, or a strong anonymization, where a few features relevant to the regular task remain. Indeed, for a significant large range of $\lambda$ values, the network randomly converges to either of these extremes, which allows only to trade-off between a few accuracies (i.e. several missing points).}
  \label{comparison}
\end{figure}

\subsection{Results analysis}

We emphasize that the present method gives an anonymizer for the whole dataset, as opposed to anonymizing a query related process or a subset of the dataset. In order to tune the  anonymization, we have trained a network for a wide range of values of $\lambda$. For each of them, we compute the accurate rates of both tasks: the private and the regular labels. 

\subsubsection*{Toggle (or sequential) vs simultaneous training} 

The procedure we found to provide better results when training the parameters of our deep neural nets is a \emph{toggle training}, as opposed to simultaneous  training ~\cite{DBLP:conf/icml/GaninL15} where all updates at the encoder and at the branches occur at the same time. With toggle training the updates are performed either at the encoder or at the branches (Figure~\ref{fig:1}). The purpose is to let the branches of the network to keep track of the encoder updates. 
This method has a key role in learning useful representations. Indeed, if classifiers are performing as efficiently as possible on their own tasks, they will feedback the most relevant information to update the encoder.
In Figure~\ref{comparison}, we confronted the result of toggled training versus the simultaneous (or concurrent) training method. The regular task accuracies are plotted as a function of the private task accuracy. To keep this comparison fair, we found better to chose a lower learning rate on the encoder than on the branches. We can observe that simultaneous training enables only two regimes: either a light anonymization, with almost no available trade-off, or a strong anonymization, where a few features relevant to the regular task remain. Indeed, after training with a significant large range of $\lambda$ values, we found the network to randomly converge to either of these extremes, that is why several points are not achievable and thus, missing in the plots.  


\begin{figure}[!]
  \centering
\includegraphics[width=13cm]{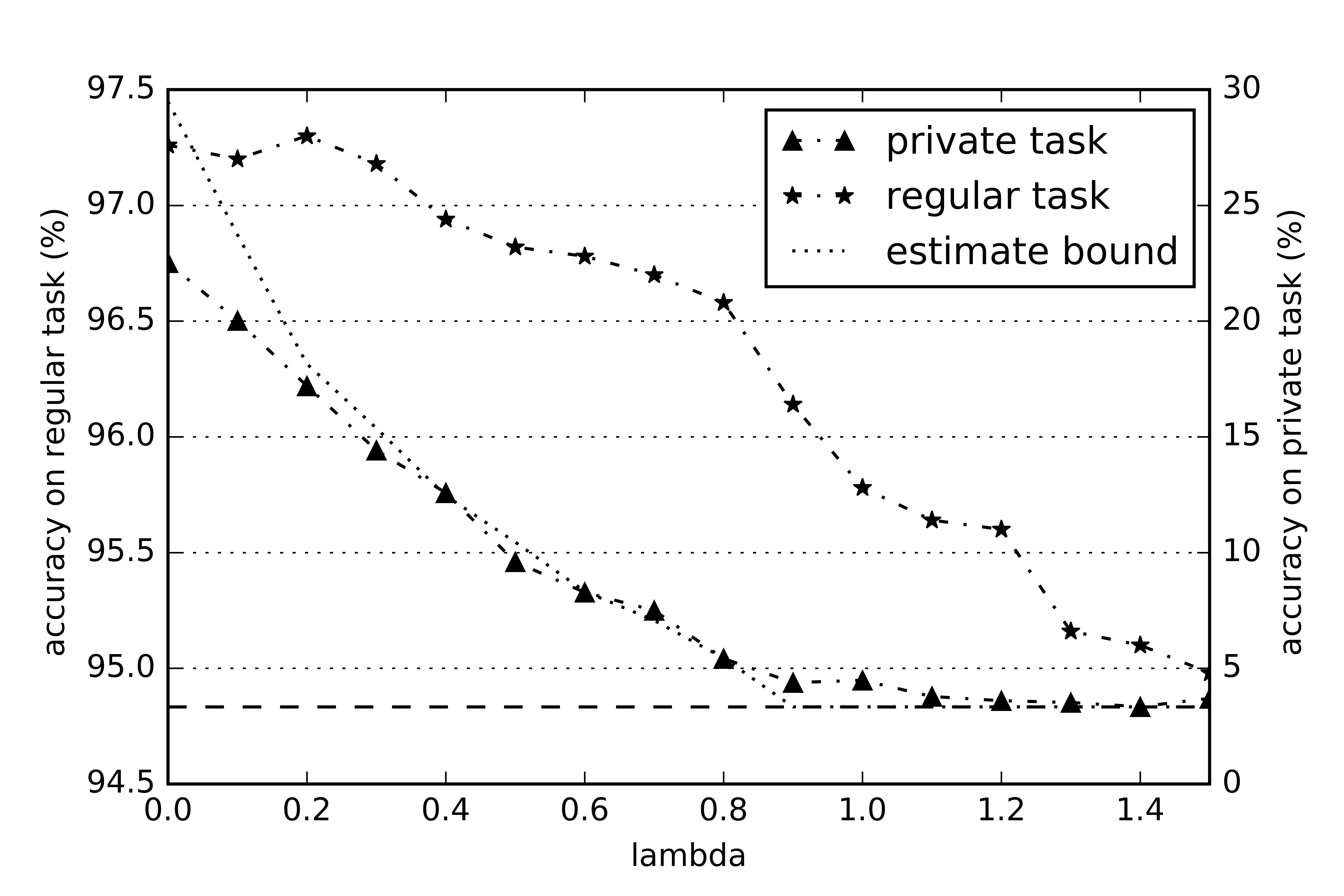} 
  \caption{Accuracies as a function of $\lambda\in[0,1.5]$ on Pen-digits database. The horizontal black dashed line is the random guessing classifier over the user-ID ($3.33\%$). 
 It displays the trade-off that occurs on the data set, i.e., a level of anonymization is ensured at the cost of a small performance decrease on the regular task. Dotes curve shows that eq.~\eqref{propo-capacity3} with \eqref{eq-approximation} is a reasonable estimation. } 
  \label{resultpendigit}
\end{figure}

\subsubsection*{Pen-digits database} 

The trade-off between these accuracies is presented in Figure~\ref{resultpendigit}. The $\blacktriangle$-curve corresponds to the test accuracy on the private task while the $\star$-curve denotes the test accuracy on the regular task. The doted curve denotes the estimation of the private task accuracy  according to ~\eqref{propo-capacity3} using \eqref{eq-approximation} computed on the loss of the test-set.  The rather good fitting indicates that \eqref{eq-approximation} is a reasonable approximation. Some interesting  conclusions  can be drawn from these plots. Its ordinate reads on the right axis.  The value of the accuracies of both tasks at $\lambda = 0$ is interesting. Indeed, when $\lambda=0$ the network is updated without any concern of the private task. On the other hand, the baseline for the private task was computed separately with a dedicated network (equivalent to cascading a network similar to the encoder and the private branch). The accuracy baseline for the private task in theses conditions was found to be around $40 \%$. Nonetheless, Figure~\ref{resultpendigit} shows a much lower accuracy because only the branch part of the network is trying to improve the classification score of the private labels, the encoder focuses in the situation of $\lambda=0$ only on the regular part. As for the regular task, it is worth to mention that the results may vary since randomness impact the training and thus the score as well. To average this noise, several simulations were made for the baseline obtaining scores between $97.65 \%$ and $98.45 \%$. The impact of $\lambda$ is quite important and is shown by the abrupt variation over the interval $\lambda\in[0,1]$. After this significant decrease in the accuracy of the private task, variations are slower, even so the accuracy of this task tends to decrease. Interestingly, regarding the score of the  regular task, variations are significantly more tenuous. Their interpretation only show that the increase in $\lambda$ does not induce any remarkable change. The impact of the private branch on the network, if such an impact exists, is rather marginal. Interestedly, the impact on the regular task  stays contained inside the previously computed baseline bound.

\begin{figure}[t]
  \centering
\includegraphics[width=13cm]{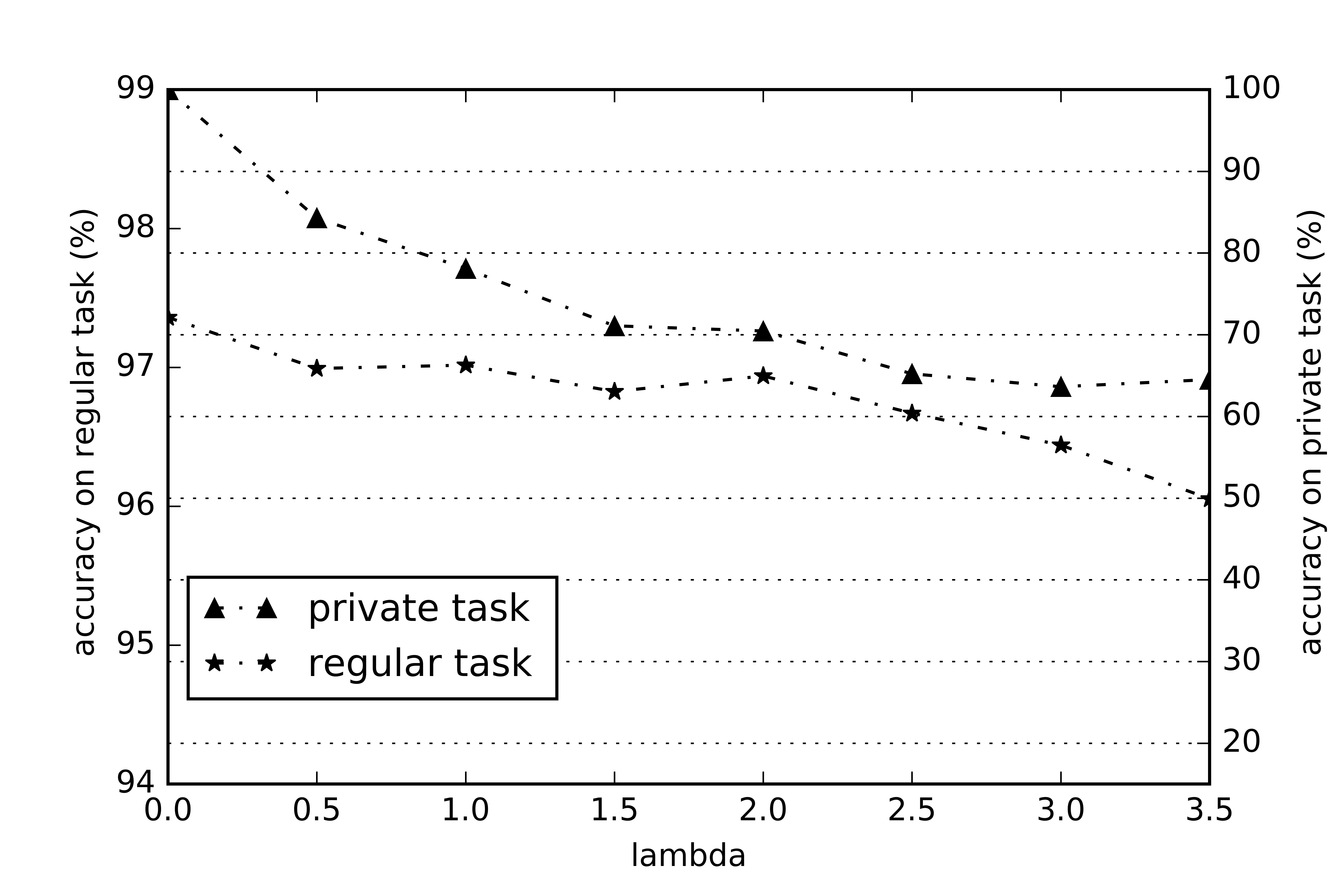} 
  \caption{Accuracies as a function of $\lambda\in[0,3.5]$ on FERG database. The horizontal black dashed line is the random guessing over the user-ID ($19.83\%$). 
The available amount of samples allow the learning of anonymized still relevant  representations but at a small cost on the regular task. For sake of clarity, $\lambda=4.5$ is not plotted since both tasks decreased to random guessing accuracy.}
  \label{fig-resultfeelFERG}
\end{figure}

\subsubsection*{FERG database} 

The plentiful samples in the database give really strong accuracies baselines for both tasks: $100 \% $ on the private task and $98.2 \%$ on the regular task. Figure~\ref{fig-resultfeelFERG} shows the trade-off, the $\star$-curve indicates  the test accuracy on the regular task which decreases from $97.36 \%$ to $96.05 \%$. The $\blacktriangle$-curve indicates the test accuracy on the private task which decreases significantly from $99.97 \%$ to $63.63 \%$. Due to the non-uniform distribution of the samples among classes, the random guessing  classifier over the user-ID is $19.83 \%$. One should notice that the six  characters have really different facial features, therefore they are easy to identify on the original images (private task baseline $100\%$). Yet, the representations learnt by the network leads to a significant anonymization with an acceptable cost on the regular task. Feeling recognition and face recognition are rather entangled tasks. The observed degradation of performance comes from the contradictory natures of both tasks, i.e. , raising the level of anonymization comes at the cost of blurring some relevant features for the regular task. Anonymization trade-offs are strongly related to the specific nature of data.

\subsubsection*{JAFFE database} 

\begin{figure}[t!]
  \centering
\includegraphics[width=13cm]{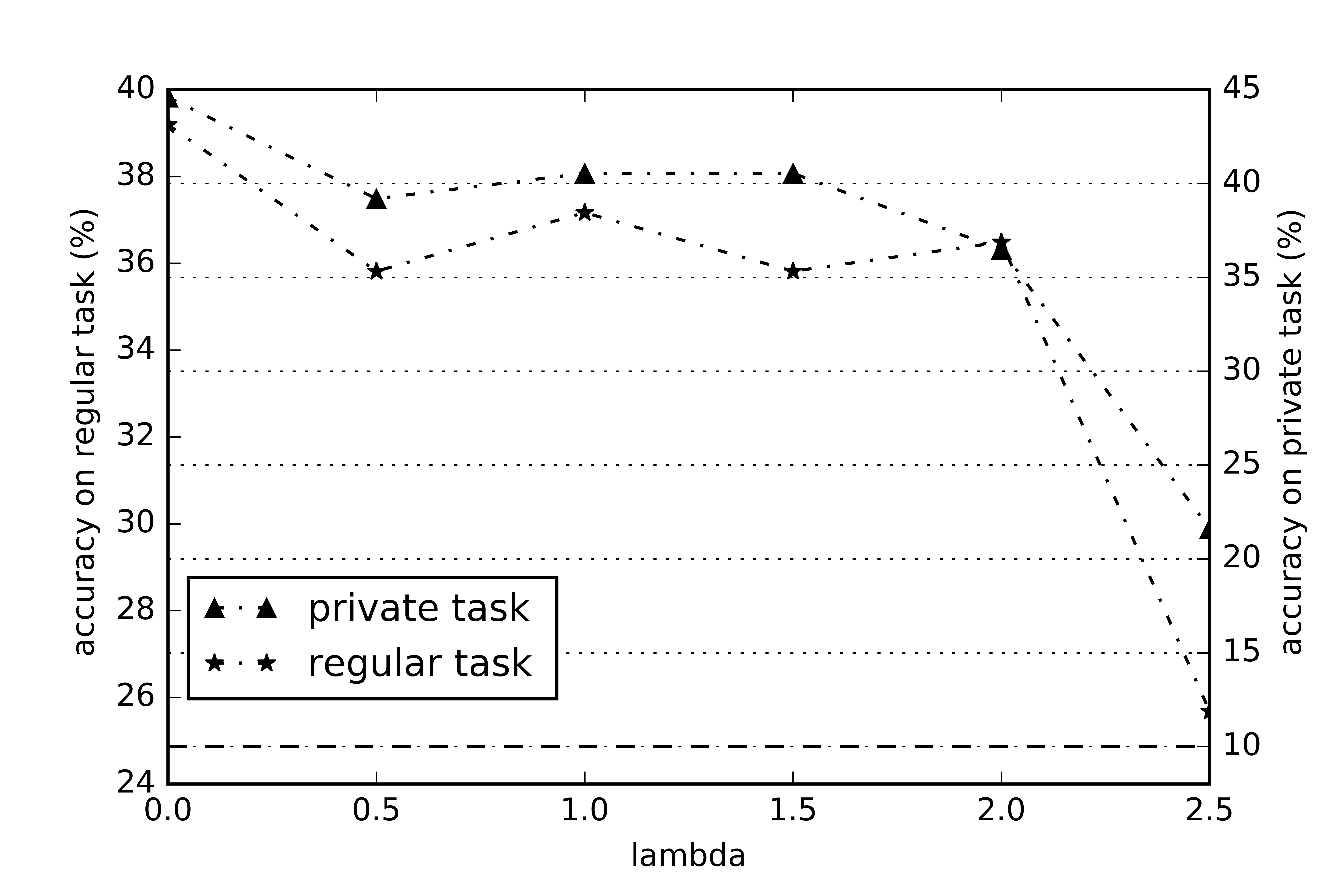} 
  \caption{Accuracies as a function of $\lambda\in[0,2.5]$ on JAFFE database. The horizontal black dashed line is the random guessing over the user-ID ($10\%$). 
  Despite a rather small dataset, the annonymization still occurs but at the cost of a significant (non-negligible) impact on the regular task performance.}
  \label{resultfeel}
\end{figure}

We note that the anonymity induced by the structure of the network itself ($\lambda=0$) is not apparent here. The accuracies of both tasks are shown in Figure~\ref{resultfeel} as a function of $\lambda$. As $\lambda$ increases,  the anonimyzation is made more apparent, i.e., the $\blacktriangle$-curve is decreasing from $44.59 \%$ to $21.62 \%$. It is clear that the trade-off is stronger on this dataset than on the previous one which can be observed from the regular task ($\star$-curve), feeling recognition, that declined from $39.19 \%$ to $25.68 \%$. This significant performance degradation is  due to the contradictory natures of the tasks but also to the limited samples, which emphasizes that encoder performance is sensitive to branches training.

\section{Summary  and Outlook}
We have presented a framework that relies information-theoretic principles to adversarial networks for learning anonymized representation of statistical data. Experimental results shown quite explicitly that a significantly large range of trade-offs can be achieved. Furthermore, the proposed method can be applied to any type of data  provided enough training samples with both regular and private labels are available, ensuring a certain trade-off between the misclassification probabilities. Extension of this work can be envisaged in many different ways but in particular, it would be important to contrast the results here to the unsupervised learning scenarios without any predefined regular task.

\section*{Acknowledgment}
We would like to acknowledge support for this project from the CNRS via the International Associated Laboratory (LIA) on \emph{Information, Learning and Control}.

\appendix
\section*{Appendix A: Proof of Lemma~\ref{lemma-surogate-representations}}

The upper bound simply follows by using Jensen-Inequality~\cite{Cover91} while the lower bound is a consequence of the definition of the rate-distortion and distortion-rate functions. The probability of misclassification  corresponding to the classifier  can be expressed in terms of the expected distortion:  
$$ 
P_{\mathcal{E}}(Q_{\hat{Z}|U},Q_{{U}|X}) = \mathbb{E}_{{P}_{XZ} {Q}_{U|X} }\left[d(Z,U) \right], 
$$ 
based on the fidelity measure $d(z,u)\coloneqq 1- Q_{\hat{Z}|U}(z|u)$. Because of the Markov chain $Z \mkv X \mkv U$, we can use the data processing inequality~\cite{Cover91} and the definition of the rate-distortion  function, obtaining the following bound for the classification error: 
\begin{align}\label{eq:boundmi}
\mathcal{I}(P_Z;Q_{U|Z}) & \geq  \min_{\rule{0mm}{4.3mm}\substack{P_{\hat{U}|Z}\,:\, \mathcal{Z}\rightarrow\mathcal{P}(\mathcal{U}) \\  \mathbb{E}_{P_{\hat{U}Z} }[d(Z,\hat{U})] \, \leq\, \mathbb{E}_{{P}_{XZ} {Q}_{U|X} }\left[d(Z,U) \right]}}\!\!\!\!\!\!\!\!\!\!\!\!\!\!\!\! \mathcal{I}\big(P_Z;P_{\hat{U}|Z}\big)\\
&= \mathcal{R}_{Z,Q_{\hat{Z}|U}}\big(P_{\mathcal{E}}(Q_{\hat{Z}|U},Q_{{U}|X}) \big).
\end{align}
For $\mathbb{E}_{{P}_{XZ} {Q}_{U|X} }\left[d(Z,U) \right]$, 
we can use the definition of $\mathcal{R}_{Z,Q_{\hat{Z}|U}}^{-1}(\cdot)$ to obtain from \eqref{eq:boundmi}, the desired inequality: 
\begin{align}
\mathcal{R}_{Z,Q_{\hat{Z}|U}}^{-1}(\mathcal{I}(P_Z;Q_{U|Z}))& \leq P_{\mathcal{E}}(Q_{\hat{Z}|U},Q_{{U}|Z}) .  
\end{align}

\bibliographystyle{apalike}
\bibliography{Clement.bib}

\end{document}